\documentclass[10pt,letterpaper]{article}

\usepackage{cogsci}

\cogscifinalcopy 

\usepackage{cmap}
\usepackage[T1]{fontenc}
\usepackage[american]{babel}
\usepackage{csquotes}
\usepackage{newtxtext,newtxmath}  
\usepackage{xcolor}     
\usepackage{tcolorbox}   
\usepackage{booktabs} 
\usepackage{caption}  
\usepackage{colortbl}

\usepackage[
  backend=biber,
  style=apa,
  natbib=true,
  annotation=false,
]{biblatex}
\usepackage{float}
\usepackage[hidelinks]{hyperref}

\title{Eliciting Trustworthiness Priors of Large Language Models via Economic Games}

\author[1]{\mbox{Siyu Yan (siyuyan02@gmail.com)\thanks{Work done while visiting the University of Hong Kong.}}}
\author[2,3,4,5,6]{\mbox{Lusha Zhu (lushazhu@pku.edu.cn)}}
\author[1]{\mbox{Jian-Qiao Zhu (zhujq@hku.hk)}}
\affil[1]{Department of Psychology, The University of Hong Kong}
\affil[2]{School of Psychological and Cognitive Sciences, Peking University}
\affil[3]{Beijing Key Laboratory of Behavior and Mental Health, Peking University}
\affil[4]{IDG/McGovern Institute for Brain Research, Peking University}
\affil[5]{Peking-Tsinghua Center for Life Sciences, Peking University}
\affil[6]{Key Laboratory of Machine Perception, Ministry of Education, China}

\begin{document}

\maketitle

\begin{abstract}
One critical aspect of building human-centered, trustworthy artificial intelligence (AI) systems is maintaining calibrated trust: appropriate reliance on AI systems outperforms both overtrust (e.g., automation bias) and undertrust (e.g., disuse). A fundamental challenge, however, is how to characterize the level of trust exhibited by an AI system itself. Here, we propose a novel elicitation method based on iterated in-context learning \citep{zhu2024eliciting} and apply it to elicit trustworthiness priors using the Trust Game from behavioral game theory. The Trust Game is particularly well suited for this purpose because it operationalizes trust as voluntary exposure to risk based on beliefs about another agent, rather than self-reported attitudes. Using our method, we elicit trustworthiness priors from several leading large language models (LLMs) and find that GPT-4.1’s trustworthiness priors closely track those observed in humans. Building on this result, we further examine how GPT-4.1 responds to different player personas in the Trust Game, providing an initial characterization of how such models differentiate trust across agent characteristics. Finally, we show that variation in elicited trustworthiness can be well predicted by a stereotype-based model grounded in perceived warmth and competence.


\end{abstract}

\section{Introduction}

With or without consciousness, delegating part or all of a task to an AI agent involves voluntary exposure to risk based on beliefs about the AI’s behavior. For example, when we ask an LLM to write a literature review, we expose ourselves to the risk of citing nonexistent papers hallucinated by the model; when we ask an LLM for investment advice, we expose ourselves to the risk that the model may be overly confident about the prospects of certain stocks. This voluntary exposure to risk, grounded in beliefs about an AI agent, aligns closely with the definition of trust in economic games \citep{berg1995trust, camerer1988experimental}. Specifically, one agent (human or artificial) invests resources (e.g., money, time, or compute) by relying on an AI agent's recommendation, while accepting the possibility of loss if the AI ``betrays'' that trust (e.g., by failing, hallucinating, or misleading).

In the context of LLMs, higher trust does not necessarily lead to better overall outcomes. The internal mechanisms of LLMs are not immediately transparent \citep{dang2024explainable}. Moreover, LLMs can be highly fluent and persuasive while remaining unreliable in certain situations, such as under distribution shift, in response to ambiguous queries, or when hallucinating \citep{bengio2025international}. LLM performance is therefore task-dependent \citep{hao2023reasoning}, uneven across cognitive domains \citep{brown2020language}, and variable across time and prompts \citep{chen2024chatgpt}. Consequently, maximizing trust risks encouraging overreliance on LLMs, whereas safe and effective collaboration with LLMs depends on maintaining appropriately calibrated levels of trust.

One key challenge, therefore, is obtaining a robust estimate of the level of trust an agent displays in multi-agent collaborations. In this paper, we focus on the classical Trust Game from behavioral game theory \citep{berg1995trust}. The Trust Game is particularly well suited to our purpose because it is a two-player cooperative game that captures trust as voluntary exposure to risk based on beliefs about the other player’s behavior. Specifically, we focus on the extent to which trust is repaid by an LLM; that is, trustworthiness. This focus reflects the structure of most human–LLM collaborations, in which humans typically act as trustors who delegate tasks to an LLM, while the LLM assumes the role of determining how much to return the trust that the human extends.

To efficiently elicit an LLM's trustworthiness prior, we adapt the idea of iterated in-context learning \citep{zhu2024eliciting} to the Trust Game by constructing a Markov chain whose samples converge to the model's prior distribution. Building on this method, we then test whether the elicited trustworthiness priors of LLMs are responsive to the persona of the trustor. Using the LLM that exhibits the most human-like trustworthiness priors, we explore the possibility of treating this model as a surrogate for a large population of human participants \citep{zhu2024incoherent, horton2023large}. Under this assumption, changes in the elicited priors of the most human-like LLM may reflect corresponding shifts in the trustworthiness judgments of the general public.

\begin{figure*}
    \centering
    \includegraphics[width=0.8\linewidth]{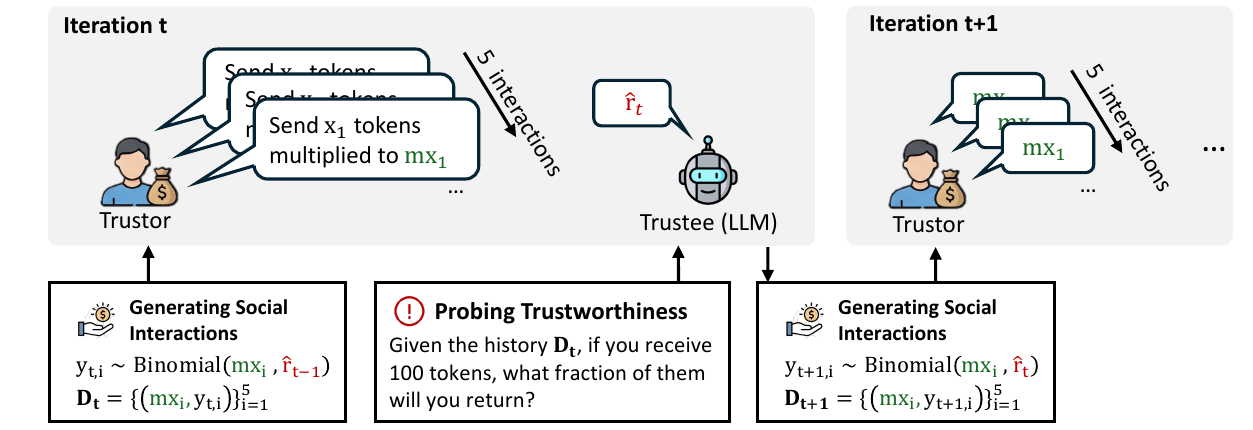}
    \caption{Illustration of the iterated in-context learning procedure used to elicit an LLM's implicit prior over trustworthiness. At iteration $t$, batch of five social interactions $\textbf{D}_t$ is generated using the previously estimated trustworthiness $\hat{r}_{t-1}$ and provided as input to the LLM, which predicts an updated trustworthiness estimate $\hat{r}_t$. This new $\hat{r}_t$ parameterizes the Binomial distribution used to generate the next iteration's batch of five social interactions, while the Trustor’s investment levels (i.e., $x_i$) remain fixed.}
    \label{fig:convergence}
\end{figure*}

\section{Background}

\textbf{Trust in AI.}
As trust can facilitate, and distrust can hinder, the adoption of advanced technologies such as AI, trust in AI has attracted significant interest among social scientists and computer scientists \citep{afroogh2024trust, liu2023trustworthy}. A central finding in this literature is that establishing trust between humans and AI systems is generally more complex than interpersonal trust \citep{thiebes2021trustworthy}. Empirically, trust in AI is distinct from trust in humans \citep{glikson2020human}. Whereas human trust involves beliefs about benevolence, integrity, and intentions, AI systems lack intentionality. As a result, trust in AI primarily concerns perceived ability and reliability, rather than honesty or goodwill. Importantly, users are not attempting to infer AI intentions, but rather to assess whether the system will behave correctly, appropriately, and dependably under uncertainty \citep{afroogh2024trust}.

\textbf{Trust Game.}
Self-reported trust (e.g., questionnaire- or Likert-scale measures) remains the dominant method for eliciting trust in AI, although behavioral measures (e.g., whether participants follow AI recommendations or cede control to AI systems) have also been used \citep{glikson2020human}. Behavioral trust can diverge from self-reported trust, as people may report distrust while still relying on AI outputs. However, most existing studies have not employed the Trust Game to decompose human–AI interactions into interpretable and incentive-compatible components. One likely reason is the assumption that AI lacks intentionality \citep{troshani2021we}. As a result, prior work typically assumes unidirectional trust, in which AI does not reciprocate in a socially meaningful way \citep{afroogh2024trust}. We argue that this assumption is increasingly outdated, as modern LLM-based AI systems exhibit sophisticated social and cognitive capabilities \citep{greenblatt2024alignment, liu2024large}. The Trust Game therefore offers an alternative bilateral framework for studying trust in human–AI interaction, in which the trustee’s responses directly reveal trustworthiness.

\textbf{Eliciting Priors using Iterated Learning.} 
Iterated learning provides a principled framework for studying how inductive biases shape behavior by examining how information is transmitted across successive generations of learners \citep{griffiths2006revealing, kalish2007iterated}. Classic work in cognitive science has shown that, when learners repeatedly infer and reproduce data generated by previous learners, the stationary distribution of the resulting transmission process converges to the learners’ prior beliefs, largely independent of the initial data distribution \citep{griffiths2007language}. As a result, iterated learning has been widely used to elicit human inductive biases in domains such as language structure, categorization, and function learning, by treating cultural transmission as a window into underlying cognitive priors \citep{griffiths2006revealing, griffiths2007language, kirby2022compositionality}. Recent work extends this framework to LLMs by leveraging in-context learning \citep{brown2020language}. \citet{zhu2024eliciting} demonstrate that iterated in-context learning can be used to elicit LLMs' implicit priors across a variety of domains of everyday cognition.


\section{Methods}

We now describe our proposed method, which integrates iterated in-context learning \citep{zhu2024eliciting} with the Trust Game \citep{berg1995trust} to elicit an LLM’s social prior over trustworthiness. We begin by formalizing the Trust Game within a Bayesian framework, in which the social prior is defined as a probability distribution over reciprocity/trustworthiness in social interaction.

\textbf{Formalizing Trust Game.}
The Trust Game is a standard paradigm in behavioral economics that formalizes social interaction between a \textit{Trustor} and a \textit{Trustee}. We further model the game within a Bayesian framework. In each interaction, both the Trustor and Trustee are endowed with an initial budget (e.g., \$10) and the Trustor chooses an amount $x \in [\$0, \$10]$ to send to the Trustee. This transferred amount is multiplied by a factor $m$ before reaching the Trustee, reflecting the potential gains from cooperation. The Trustee then decides how much of the received amount to return to the Trustor. Let $y \in \{\$0, ~\$1, ~\$2, \ldots, ~\$mx\}$ denote the returned amount.

To express trustworthiness independently of the scale of the investment, we define the \textit{return ratio}:
\begin{align}
    r = \frac{y}{mx} \in [0, 1]
\end{align}
which represents the fraction of the available resources that the Trustee chooses to return. Higher values of $r$ correspond to greater trustworthiness.

We assume that, conditional on a latent trustworthiness parameter $r$, the returned amount follows a Binomial likelihood,
\begin{align}
    p(y|x,r) = \text{Binomial}(y | mx, r)
\end{align}
where $r$ is interpreted as the probability that any unit of the multiplied investment is returned. Prior beliefs about trustworthiness are captured by a Beta prior,
\begin{align}
    p(r) = \text{Beta}(r | \alpha, \beta)
\end{align}
with shape parameters $\alpha$ and $\beta$. Together, the Binomial likelihood and Beta prior define a Beta–Binomial model of reciprocity in the Trust Game. In this formulation, the prior distribution $p(r)$ represents the Trustee’s a prior expectations about how much trust is typically repaid. In this work, $p(r)$ is the target prior distribution that we seek to elicit from LLM, as it directly reflects their inductive biases about trustworthiness in social exchange.

\textbf{Recovering Trustworthiness Priors via Iterated Learning.} 
The Bayesian framing of the Trust Game naturally induces an iterated learning process that can converge to a model’s trustworthiness prior, $p(r)$. Specifically, this process can be viewed as simulating a Markov chain over prompts, in which the model repeatedly observes previous social interactions from the game and predicts the Trustee’s behavior.

Critically, we impose an \textit{information bottleneck}: at each iteration $t$, the model observes only a limited batch of $B$ previous interactions, denoted by
\begin{align}
    \mathbf{D}_{t-1} = \{(mx_{t-1,i}, y_{t-1,i})\}_{i=1}^{B}
\end{align}

For a Bayesian agent under the Beta–Binomial model, the posterior distribution over trustworthiness takes a closed-form Beta distribution:
\begin{align}
    p(r|\mathbf{D}_{t-1}) = \text{Beta} \Big(\underbrace{\alpha+\sum_{i=1}^B {y_{t-1,i}}}_{\alpha'}, ~\underbrace{\beta+\sum_{i=1}^B (mx_{t-1,i}-y_{t-1,i})}_{\beta'} \Big)
    \label{eq:bayes_posterior}
\end{align}
where the prior shape parameters $\alpha$ and $\beta$ are updated to $\alpha'$ and $\beta'$ based on the observed social interactions. The imposed information bottleneck ensures that only a limited amount of evidence is retained at each iteration, causing information to decay across iterations. As a result, the iterated learning process reveals the model’s underlying inductive biases rather than allowing it to accumulate or memorize past observations.

As illustrated in Figure \ref{fig:convergence}, at each iteration of the iterated in-context learning, the LLM is prompted with the dataset $\mathbf{D}_{t-1}$ and asked to infer the posterior mean of the trustworthiness parameter, denoted $\hat{r}$. To generate the social interaction data for the next iteration, we treat $\hat{r}$ as the model’s estimate of reciprocity and simulate a new batch of interactions $\mathbf{D}_{t}=\{(mx_{t,i}, y_{t,i})\}_{i=1}^{B}$, where each returned amount is sampled according to the likelihood function $y_{t,i} \sim \text{Binomial}(mx_{t,i}, \hat{r})$.

This method closely parallels iterated learning with coin flips, where learners repeatedly infer a coin's bias from a small sample of flips and then generate new flips based on their updated beliefs about the coin's bias. In such settings, it is well established that the distribution of inferred biases converges to the learner’s prior over coin biases, regardless of the initial data \citep{zhu2024eliciting, reali2009evolution}. In our formulation, trustworthiness $r$ plays the role of the coin’s bias, with returned and unreturned units corresponding to successes and failures, respectively. Iterated learning in the Trust Game therefore recovers the model’s implicit prior $p(r)$ over reciprocity/trustworthiness through the same dynamics.

\textbf{Implementational details.}
In implementing our method, we set the size of the transmitted dataset of social interactions to $B=5$, meaning that only five interactions are passed between successive generations of LLMs. We further fixed the Trustor’s investment to five values that span the range of possible investments, $x_{t,i}\in\{\$1, ~\$3, ~\$5, ~\$7, ~\$9\}$, out of a \$10 endowment, for all iterations $t$. Finally, we fixed the multiplication factor applied to the Trustor’s investment at $m=3$. Consequently, only the amounts returned by the Trustee (i.e., $y_{t,i}$) vary across iterations, and these returns are determined by the estimated trustworthiness $\hat{r}$ from the previous iteration.

\begin{figure*}
    \centering
     \includegraphics[width=0.86\linewidth]{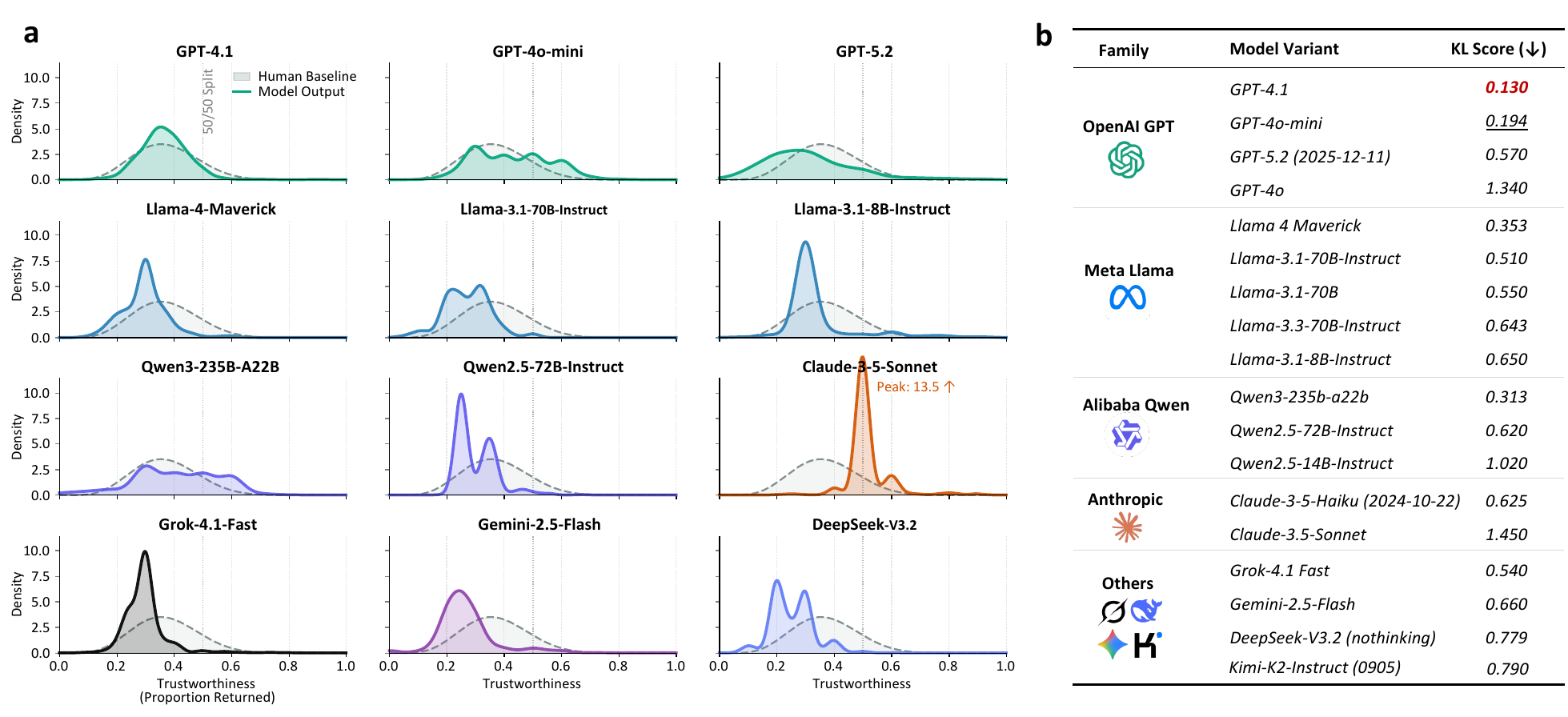}
    \caption{\textbf{(a)} Elicited trustworthiness priors for a range of LLMs (model versions shown as titles in each panel). Human trustworthiness distributions, adapted from the meta-analysis by \citet{johnson2011trust}, are overlaid in each panel as semi-transparent grey distributions for comparison. The human mean return ratio is 0.372 with a standard deviation of 0.114. All experiments were conducted between November 2025 and January 2026. \textbf{(b)} KL divergence by model family: $D_{KL}(p_\text{LLM}(r) \parallel p_\text{human}(r))$. Lower values indicate elicited trustworthiness distributions that are closer to the human baseline.}
    \label{fig:Alignment}
\end{figure*}

\section{Experiment 1: Eliciting Trustworthiness Priors}


\textbf{Models.}
We evaluated a comprehensive set of 20 LLMs, comprising both leading proprietary models and state-of-the-art open-weight models, to provide a representative benchmark. Specifically, the proprietary suite included models from the \textit{GPT-4} and \textit{GPT-5} families, the \textit{Claude-3.5} series, \textit{Gemini-2.5-Flash}, and \textit{Grok-4.1-Fast}. The open-weight suite covered a diverse range of architectures, including the \textit{Llama-3.x} series (8B to 70B), \textit{Llama-4-Maverick}, the \textit{Qwen-2.5} family (14B to 72B), \textit{Qwen-3-235B} (22B active parameters), as well as specialized architectures such as \textit{DeepSeek-V3.2} and \textit{Kimi-K2-Instruct} (see Figure \ref{fig:Alignment}b).

For all models, we set the temperature to 1 to sample directly from the model logits. The iterated learning process was initialized with nine seed values of trustworthiness, $r\in\{ 0.1, 0.2, \ldots, 0.9\}$. Each chain was run for 30 iterations and independently re-initialized 30 times for each seed, yielding a total of 270 chains per model.

\textbf{Prompt Design.}
Inspired by \citet{zhu2024eliciting}, we designed the prompts, and explicitly asked the model to provide rapid responses.
At each iteration, the five most recent social interactions were passed to the model via the placeholder $\texttt{\{past\_social\_interactions\}}$, which contained sampled outcomes generated from the Binomial distribution (see Figure~\ref{fig:convergence}). The template prompt used to elicit the trustworthiness prior is shown below:

``\textit{Please imagine that you are a behavioral economics researcher. You are investigating an experiment involving two isolated groups: `Player A' and `Player B'. In this experiment, player A and player B both receive an initial endowment from the experimenter. Player A can send $N$ dollars out of it, and player B will receive $3N$ dollars. The player B then decides how much to return to the player A. You will see some information about the decisions made by previous pairs. Within the previous 5 pairs, the following transactions occurred: \texttt{\{past\_social\_interactions\}} (Note: $m_x =$ amount player B received; $y =$ amount player B returned). Based on the past interactions, suppose the player B receives a total of 100 tokens in a new interaction. What fraction of the received funds do you predict he/she will return to the player A? Please limit your answer to a single value between 0 and 1, without outputting anything else.}''

\textbf{Results.}
To assess convergence of the iterated-learning chains, we applied the Gelman–Rubin diagnostic ($\hat{R}$) \citep{gelman1992inference}, which quantifies the ratio of within-chain variance to between-chain variance. We adopted the standard threshold of $\hat{R} \le 1.1$ as an indicator of satisfactory convergence. All models exhibited rapid convergence, typically reaching stability within 20 iterations--well below the maximum chain length of 30 iterations.



To assess whether the elicited priors capture the inductive biases that guide LLMs’ social decisions in the Trust Game, we elicited an additional set of single-shot behaviors from the models. We used the same Trust Game framework as in the iterated in-context learning procedure, but reduced the number of observed past social interactions from $B=5$ to a single observation. We fixed the multiplier $m$ at 3 and varied the investment level $x$ across all integer values between $[0,8]$. For each condition, the LLM assumes the role of the Trustee by specifying a returned amount $y$, which serves as the target behavior for the Bayesian model to predict.

To account for LLMs’ reciprocity behavior, we used the Bayesian model in Equation~\ref{eq:bayes_posterior} to evaluate three alternative priors: (1) a uniform prior, (2) a human prior derived from a meta-analysis of Trust Games \citep{johnson2011trust}, and (3) the elicited prior obtained via iterated in-context learning. For all three cases, we employed numerical methods by discretizing the parameter space of $r$ into a grid of 101 evenly spaced points over $[0,1]$. The mean of the resulting posterior distribution was taken as the model’s prediction of the LLM’s reciprocity behavior.

As shown in Table~\ref{tab:prior_comparison}, the elicited prior outperforms both the uniform and human priors when evaluated using root mean squared deviation (RMSD) and Pearson’s $r$. These results suggest that the iterated in-context learning method successfully recovers the LLMs’ implicit prior over trustworthiness that can better predict their own reciprocity behavior.

\begin{table}[t!] 
\centering
\caption{Comparison of Bayesian models of reciprocity behavior with various priors and \textit{GPT-4.1}’s decisions using Pearson’s $r$ and root-mean-squared deviation (RMSD).}
\label{tab:prior_comparison}
\small 
\begin{tabular}{lcc}
\toprule
\textbf{Priors} & \textbf{RMSD} ($\downarrow$) & \textbf{Pearson's $r$} ($\uparrow$) \\ \midrule
Uniform Prior         & 0.1609 & 0.7938 \\
Human Prior    & 0.1270 & 0.8165 \\ 
Elicited Prior           & \textbf{0.1190} & \textbf{0.8225} \\\bottomrule
\end{tabular}
\end{table}

After confirming convergence of the iterated learning chains and that the elicited priors effectively predict LLMs’ reciprocity behavior, we examined the alignment between the elicited trustworthiness priors of LLMs and empirical human priors derived from a meta-analysis of Trust Games played between humans \citep{johnson2011trust}, as shown in Figure~\ref{fig:Alignment}a. To quantify the (dis)similarity between the LLM-elicited priors and human trustworthiness priors, we computed the Kullback–Leibler divergence, $D_{KL}(p_\text{LLM}(r) \parallel p_\text{human}(r))$. As summarized in Figure~\ref{fig:Alignment}b, we find that different LLMs exhibit varying degrees of correspondence with the human trustworthiness distribution, with GPT-4.1 showing the strongest correspondence, as indicated by the smallest KL divergence ($D_{KL}(p_\text{GPT-4.1}(r) \parallel p_\text{human}(r))=0.130$).

Interestingly, some models, such as \textit{Qwen2.5-72B} or \textit{Llama-3.1-70B-Instruct}, exhibit bimodal trustworthiness priors, suggesting a mixture of strategic responses corresponding to different trust behaviors. In addition, several models, including \textit{Grok-4.1-Fast} and \textit{GPT-5.2}, display a systematic downward shift in trustworthiness priors (toward mean $r<0.372$), indicating lower average levels of trust relative to human baselines.

\begin{figure}[t]
    \centering
    \includegraphics[width=1\linewidth]{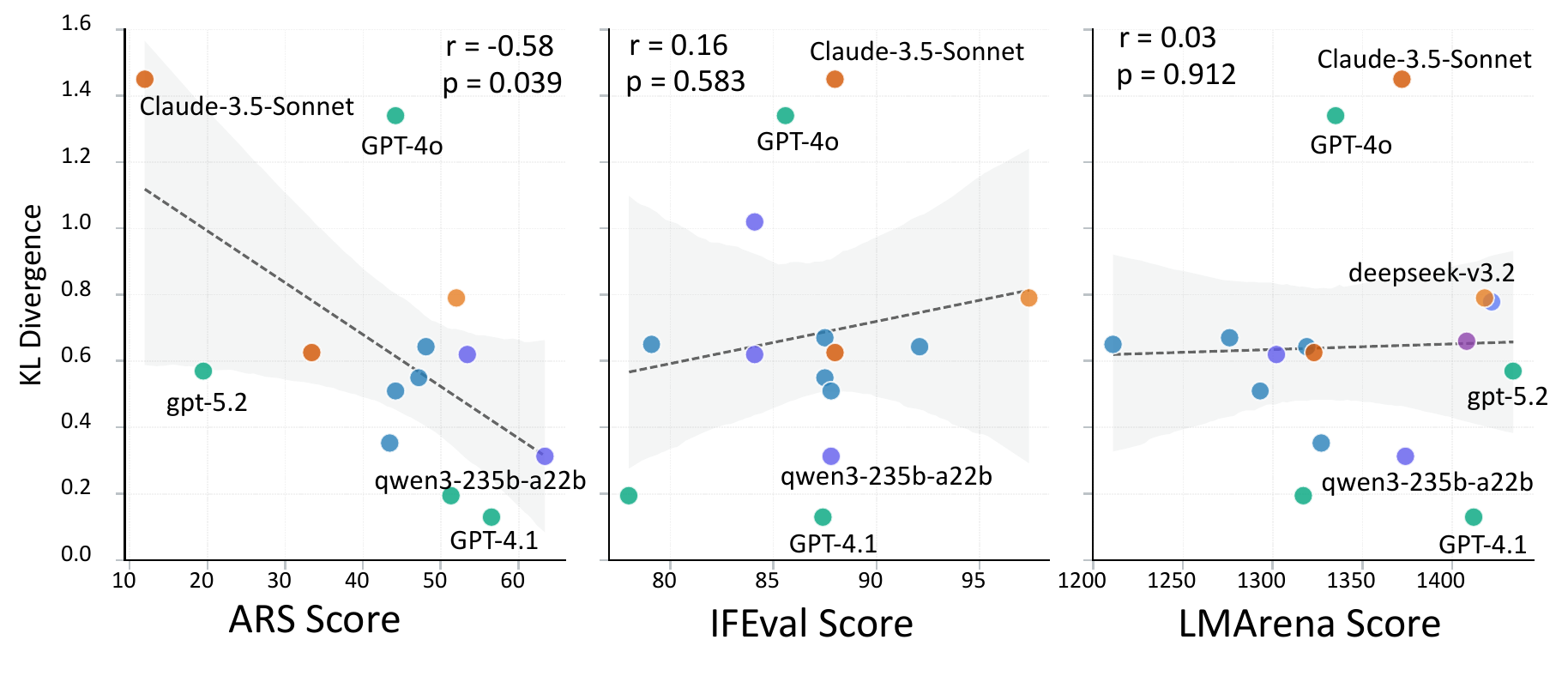}
    \caption{Correlation analysis between KL Divergence value and model performance metrics using Pearson’s r. Panels (left to right) correspond to Average Risk Score (ARS) for safety and risk propensity, IFEval for instruction-following strictness, and LMArena for general reasoning capability.}
    \label{fig:model_attributes}
\end{figure}


Given the diversity of trustworthiness priors elicited across different LLMs, we next examine whether the degree of alignment between LLM and human priors correlates with model capabilities, as measured by benchmarks that involve social or interactive behaviors. We focus on three such benchmarks on which all evaluated LLMs have been tested: (1) the Average Risk Score (ARS) from the FORTRESS benchmark \citep{knight2025fortress}\footnote{\url{https://scale.com/leaderboard/fortress}}, (2) the Instruction-Following Evaluation (IFEval) \citep{zhou2023instruction}, and (3) LMArena \footnote{\url{https://lmarena.ai/}} scores collected between January 9 and 13, 2026. 

ARS quantifies the expected risk level of LLM outputs across a curated set of safety-relevant scenarios, with lower scores indicating fewer or less severe risky outputs. IFEval measures how well a model follows explicit instructions, particularly when those instructions involve strict constraints such as required formats, item counts, or structural rules. Finally, LMArena evaluates human preferences through blind, head-to-head comparisons between model outputs, aggregating these judgments into an Elo-style score. Combined, these benchmarks capture complementary aspects of human–AI interaction that may relate to voluntary exposure to risk, as operationalized in the Trust Game setting.

We performed a correlation analysis between these benchmark scores and the dissimilarity between LLM-elicited and human trustworthiness priors; the results are shown in Figure~\ref{fig:model_attributes}. Notably, we observe that models with higher risk propensity, as indicated by higher ARS scores, tend to exhibit greater alignment with human trustworthiness priors ($r=-0.58, p=0.039$). In contrast, alignment with human priors shows no significant correlation with either LM Arena scores or instruction-following performance as measured by IFEval. Overall, while the association with ARS suggests a potential link between risk-related behavior and trustworthiness priors, no broad or robust correlational patterns emerge across the evaluated benchmarks.



 



\section{Experiment 2: Trustworthiness Priors Across Trustor Personas}

Given the strong alignment between the elicited trustworthiness priors of \textit{GPT-4.1} and those observed in humans, we hypothesize that this model can serve as the best-performing simulator of human behavior in the Trust Game among the LLMs evaluated. Under this assumption, we use \textit{GPT-4.1} to simulate social interactions with Trustors characterized by different personas. Specifically, \textit{GPT-4.1} is treated as a surrogate human Trustee to estimate how an average human player might behave in the Trust Game when interacting with Trustors possessing different characteristics. 

\begin{figure}[t!]
    \centering
    \includegraphics[width=0.9\linewidth]{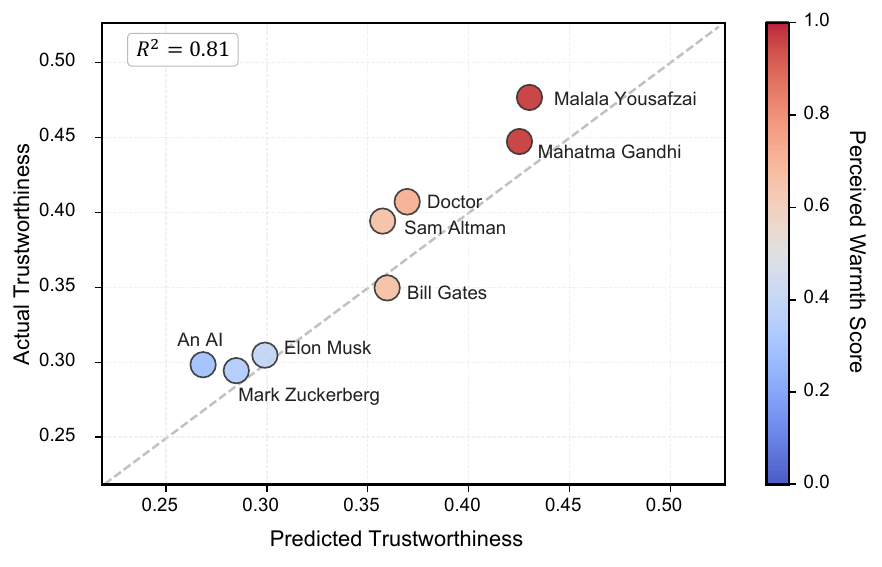}
    \caption{Stereotype-based models grounded in warmth and competence effectively predict \textit{GPT-4.1}’s mean elicited trustworthiness across different Trustor personas. The horizontal axis shows trustworthiness predicted by the regression model reported in Table~\ref{tab:regression_results}, while the vertical axis shows trustworthiness elicited via iterated in-context learning.}
    \label{fig:persona}
\end{figure}

\textbf{Trustor Personas.} We replaced the neutral description of the Trustor (i.e., ``player 1'') with a set of more specific persona descriptions, including occupational roles (e.g., an AI, a doctor) as well as well-known public figures (e.g., Elon Musk, Mark Zuckerberg, Bill Gates, Sam Altman, Mahatma Gandhi, and Malala Yousafzai) (see Figure \ref{fig:persona}). The iterated in-context learning procedure was then used to generate trustworthiness predictions for each persona by prompting \textit{GPT-4.1} with the corresponding persona descriptions.

\textbf{Results.} The elicited trustworthiness priors of \textit{GPT-4.1} are responsive to Trustor personas, exhibiting systematic variation across different persona descriptions (mean trustworthiness ranged from 0.298 to 0.447). This variation may reflect corresponding differences in human trustworthiness judgments toward these personas (e.g., people may place greater trust in Malala Yousafzai than in Mark Zuckerberg).

\textbf{Explaining variation in trustworthiness.}
We next seek to explain the systematic variation observed in trustworthiness across Trustor personas. A prominent account proposed by \citet{fiske2018stereotype} explains social behaviors in terms of stereotypes organized along two fundamental dimensions: warmth (perceived intentions) and competence (perceived ability). Motivated by computational framework established by \citet{jenkins2018predicting}, we construct a simple linear model of trustworthiness as a function of warmth and competence. Specifically, the model regresses trustworthiness on warmth, competence, and their interaction.

\begin{table}[t!]
\centering
\caption{Estimated main and interaction effects for return rate.}
\label{tab:regression_results}
\small 
\setlength{\tabcolsep}{3pt} 
\begin{tabular}{l cccc} 
\toprule
 & & \multicolumn{2}{c}{Main Effects} & Interaction \\
\cmidrule(lr){3-4} \cmidrule(lr){5-5}
 Variable & Intercept &  $W$ &  $C$ & $W \times C$ \\
\midrule
Trustworthiness & $0.184^{***}$ & $0.161^{***}$ & $0.017^{***}$ & $0.086^{***}$ \\
 & (0.003) & (0.004) & (0.004) & (0.005) \\
\bottomrule
\multicolumn{5}{l}{\footnotesize \textit{Note:} Standard errors are in parentheses. $^{***}p < 0.001$.}
\end{tabular}
\end{table}

To estimate the coefficients of this linear model predicting trustworthiness from warmth and competence, we employ a $2\times 2$ factorial design. Specifically, we elicit \textit{GPT-4.1}’s trustworthiness priors for four Trustor personas that independently vary in warmth and competence, each at high or low levels. For example, a persona high in both warmth and competence is described as ``\textit{a highly respected expert in their field, widely celebrated for their exceptional professional achievements and intelligence. Simultaneously, they are known for their profound benevolence, mentorship, and active care for others}''. We then regress the elicited trustworthiness priors on dummy variables representing warmth ($W$) and competence ($C$) using ordinary least squares:
\begin{align}
r = \beta_0 + \beta_1 W + \beta_2 C + \beta_3 (W \times C) + \epsilon
\end{align}
where $W$ and $C$ are dummy variables ($0 = \text{Low}, 1 = \text{High}$), and $\epsilon$ denotes the Gaussian error term. 

Table~\ref{tab:regression_results} reports the regression results. The findings are consistent with the primacy of warmth hypothesis commonly observed in human social psychology \citep{fiske2007universal}. The intercept ($\beta_0 = 0.184$) reflects a baseline tendency toward resource preservation in the absence of positive social signals. Warmth exerts a substantially stronger influence than competence ($\beta_1 \gg \beta_2$): benevolence alone leads to a marked increase in trustworthiness, whereas competence in the absence of warmth yields only a negligible marginal gain. Moreover, the significant interaction term ($\beta_{3}$) indicates a synergistic effect, whereby competence contributes meaningfully to trustworthiness primarily when paired with warmth. This pattern suggests a significant interaction effect of social features, such that agents perceived as both high in warmth and competence are valued well beyond the additive contributions of each trait alone.


\textbf{Generalizing to realistic personas.}
Having obtained the regression model grounded in stereotype theory that predicts \textit{GPT-4.1}’s trustworthiness from perceived warmth and competence, we next examine whether this regression model generalizes to more realistic Trustor personas. To this end, we first prompted \textit{GPT-4.1} to provide numerical ratings of warmth and competence for the public figures and generic personas shown in Figure~\ref{fig:persona}. The model was instructed to rate each persona’s warmth and competence on a continuous scale from 0 to 1. For example, Elon Musk was rated as having warmth and competence values of 0.40 and 0.99, respectively, whereas Mahatma Gandhi was rated as having warmth and competence values of 0.95 and 0.90.

Using these warmth and competence scores, we applied the regression model reported in Table~\ref{tab:regression_results} to predict the trustworthiness of each Trustor persona. Because we also independently elicited trustworthiness for each persona using iterated in-context learning, we were able to assess how closely the regression model’s predictions aligned with the elicited trustworthiness values. As shown in Figure~\ref{fig:persona}, the regression model captures a substantial proportion of the variance in elicited trustworthiness using only warmth and competence scores ($R^2=0.81$). This result suggests that these two stereotypical dimensions can be effectively leveraged to predict an LLM’s trustworthiness across different Trustor personas.

\section{Discussion}

We extend the iterated in-context learning method to the trust game and elicit trustworthiness priors from a range of LLMs. Applying the same procedure reveals substantial variation in the elicited trustworthiness priors across models. We then focus our analysis on the LLM that exhibits the most human-like trustworthiness prior, \textit{GPT-4.1}, and prompt this model to play the game with Trustors characterized by different personas. We observe systematic variation in the elicited trustworthiness priors across Trustor personas. Moreover, these variations are well explained by a stereotype-based model that accounts for the Trustor’s perceived warmth and competence.

\textbf{Limitations and Future Research.}
One limitation of our proposed method is the assumption that the trustworthiness prior is stable across social interactions. This need not be the case: if the trustworthiness parameter $r$ varies with the investment level $x$ or across interactions, a hierarchical or conditional model of reciprocity would be more appropriate. Recent work on human strategic choice suggests that context-dependent strategic behavior provides a better account of human decision-making \citep{zhu2025capturing}, and such context sensitivity is likely to be shared by LLMs that are aligned with human preferences.

Finally, while access to the internal weights of proprietary models is generally limited to researchers within AI companies, open-weight models provide a viable avenue for investigating the neural representations of trustworthiness in LLMs. Moreover, understanding these neural representations may inform the design and control of more trustworthy AI systems.

\printbibliography

@article{zhou2023instruction,
  title={Instruction-following evaluation for large language models},
  author={Zhou, Jeffrey and Lu, Tianjian and Mishra, Swaroop and Brahma, Siddhartha and Basu, Sujoy and Luan, Yi and Zhou, Denny and Hou, Le},
  journal={arXiv preprint arXiv:2311.07911},
  year={2023}
}

@article{camerer1988experimental,
  title={Experimental tests of a sequential equilibrium reputation model},
  author={Camerer, Colin and Weigelt, Keith},
  journal={Econometrica: Journal of the Econometric Society},
  pages={1--36},
  year={1988},
  publisher={JSTOR}
}

@article{zhu2025capturing,
  title={Capturing the complexity of human strategic decision-making with machine learning},
  author={Zhu, Jian-Qiao and Peterson, Joshua C and Enke, Benjamin and Griffiths, Thomas L},
  journal={Nature Human Behaviour},
  pages={1--7},
  year={2025},
  publisher={Nature Publishing Group UK London}
}

@article{reali2009evolution,
  title={The evolution of frequency distributions: Relating regularization to inductive biases through iterated learning},
  author={Reali, Florencia and Griffiths, Thomas L},
  journal={Cognition},
  volume={111},
  number={3},
  pages={317--328},
  year={2009},
  publisher={Elsevier}
}

@inproceedings{kirby2022compositionality,
  title={Compositionality Arises from Iterated Learning Despite a Preference for Holistic Signals: An Experimental Model of Sign Language Emergence’},
  author={Kirby, SIMON and Ravignani, A},
  booktitle={The Evolution of Language: Proceedings of the Joint Conference on Language Evolution (JCoLE)},
  pages={402--4},
  year={2022},
  organization={Joint Conference on Language Evolution Nijmegen}
}

@article{kalish2007iterated,
  title={Iterated learning: Intergenerational knowledge transmission reveals inductive biases},
  author={Kalish, Michael L and Griffiths, Thomas L and Lewandowsky, Stephan},
  journal={Psychonomic bulletin \& review},
  volume={14},
  number={2},
  pages={288--294},
  year={2007},
  publisher={Springer}
}

@article{troshani2021we,
  title={Do we trust in {AI}? Role of anthropomorphism and intelligence},
  author={Troshani, Indrit and Rao Hill, Sally and Sherman, Claire and Arthur, Damien},
  journal={Journal of Computer Information Systems},
  volume={61},
  number={5},
  pages={481--491},
  year={2021},
  publisher={Taylor \& Francis}
}

@article{liu2023trustworthy,
  title={Trustworthy llms: a survey and guideline for evaluating large language models' alignment},
  author={Liu, Yang and Yao, Yuanshun and Ton, Jean-Francois and Zhang, Xiaoying and Guo, Ruocheng and Cheng, Hao and Klochkov, Yegor and Taufiq, Muhammad Faaiz and Li, Hang},
  journal={arXiv preprint arXiv:2308.05374},
  year={2023}
}

@article{liu2024large,
  title={Large language models assume people are more rational than we really are},
  author={Liu, Ryan and Geng, Jiayi and Peterson, Joshua C and Sucholutsky, Ilia and Griffiths, Thomas L},
  journal={arXiv preprint arXiv:2406.17055},
  year={2024}
}

@article{greenblatt2024alignment,
  title={Alignment faking in large language models},
  author={Greenblatt, Ryan and Denison, Carson and Wright, Benjamin and Roger, Fabien and MacDiarmid, Monte and Marks, Sam and Treutlein, Johannes and Belonax, Tim and Chen, Jack and Duvenaud, David and others},
  journal={arXiv preprint arXiv:2412.14093},
  year={2024}
}

@article{glikson2020human,
  title={Human trust in artificial intelligence: Review of empirical research},
  author={Glikson, Ella and Woolley, Anita Williams},
  journal={Academy of management annals},
  volume={14},
  number={2},
  pages={627--660},
  year={2020},
  publisher={Briarcliff Manor, NY}
}

@article{thiebes2021trustworthy,
  title={Trustworthy artificial intelligence},
  author={Thiebes, Scott and Lins, Sebastian and Sunyaev, Ali},
  journal={Electronic Markets},
  volume={31},
  number={2},
  pages={447--464},
  year={2021},
  publisher={Springer}
}

@article{afroogh2024trust,
  title={Trust in {AI}: progress, challenges, and future directions},
  author={Afroogh, Saleh and Akbari, Ali and Malone, Emmie and Kargar, Mohammadali and Alambeigi, Hananeh},
  journal={Humanities and Social Sciences Communications},
  volume={11},
  number={1},
  pages={1--30},
  year={2024},
  publisher={Palgrave}
}

@article{bengio2025international,
  title={International {AI} Safety Report 2025: First Key Update: Capabilities and Risk Implications},
  author={Bengio, Yoshua and Clare, Stephen and Prunkl, Carina and Rismani, Shalaleh and Andriushchenko, Maksym and Bucknall, Ben and Fox, Philip and Hu, Tiancheng and Jones, Cameron and Manning, Sam and others},
  journal={arXiv preprint arXiv:2510.13653},
  year={2025}
}

@article{zhu2024incoherent,
  title={Incoherent probability judgments in large language models},
  author={Zhu, Jian-Qiao and Griffiths, Thomas L},
  journal={Annual Meetings of the Cognitive Science Society},
  year={2024}
}

@techreport{horton2023large,
  title={Large language models as simulated economic agents: What can we learn from homo silicus?},
  author={Horton, John J},
  year={2023},
  institution={National Bureau of Economic Research}
}

@article{dang2024explainable,
  title={Explainable and interpretable multimodal large language models: A comprehensive survey},
  author={Dang, Yunkai and Huang, Kaichen and Huo, Jiahao and Yan, Yibo and Huang, Sirui and Liu, Dongrui and Gao, Mengxi and Zhang, Jie and Qian, Chen and Wang, Kun and others},
  journal={arXiv preprint arXiv:2412.02104},
  year={2024}
}

@article{chen2024chatgpt,
  title={How is ChatGPT’s behavior changing over time?},
  author={Chen, Lingjiao and Zaharia, Matei and Zou, James},
  journal={Harvard Data Science Review},
  volume={6},
  number={2},
  year={2024},
  publisher={The MIT Press}
}

@article{brown2020language,
  title={Language models are few-shot learners},
  author={Brown, Tom and Mann, Benjamin and Ryder, Nick and Subbiah, Melanie and Kaplan, Jared D and Dhariwal, Prafulla and Neelakantan, Arvind and Shyam, Pranav and Sastry, Girish and Askell, Amanda and others},
  journal={Advances in neural information processing systems},
  volume={33},
  pages={1877--1901},
  year={2020}
}

@inproceedings{hao2023reasoning,
  title={Reasoning with language model is planning with world model},
  author={Hao, Shibo and Gu, Yi and Ma, Haodi and Hong, Joshua and Wang, Zhen and Wang, Daisy and Hu, Zhiting},
  booktitle={Proceedings of the 2023 Conference on Empirical Methods in Natural Language Processing},
  pages={8154--8173},
  year={2023}
}

@article{gelman1992inference,
  title={Inference from iterative simulation using multiple sequences},
  author={Gelman, Andrew and Rubin, Donald B},
  journal={Statistical science},
  volume={7},
  number={4},
  pages={457--472},
  year={1992},
  publisher={Institute of Mathematical Statistics}
}

@article{berg1995trust,
  title={Trust, reciprocity, and social history},
  author={Berg, Joyce and Dickhaut, John and McCabe, Kevin},
  journal={Games and economic behavior},
  volume={10},
  number={1},
  pages={122--142},
  year={1995},
  publisher={Elsevier}
}

@article{zhu2024eliciting,
  title={Eliciting the priors of large language models using iterated in-context learning},
  author={Zhu, Jian-Qiao and Griffiths, Thomas L},
  journal={Annual Meetings of the Cognitive Science Society},
  year={2024}
}

@article{griffiths2007language,
  title={Language evolution by iterated learning with Bayesian agents},
  author={Griffiths, Thomas L and Kalish, Michael L},
  journal={Cognitive science},
  volume={31},
  number={3},
  pages={441--480},
  year={2007},
  publisher={Wiley Online Library}
}

@inproceedings{griffiths2006revealing,
  title={Revealing priors on category structures through iterated learning},
  author={Griffiths, Thomas L and Christian, Brian R and Kalish, Michael L},
  booktitle={Proceedings of the 28th annual conference of the Cognitive Science Society},
  volume={199},
  year={2006}
}

@article{knight2025fortress,
  title={FORTRESS: Frontier Risk Evaluation for National Security and Public Safety},
  author={Knight, Christina Q and Deshpande, Kaustubh and Sirdeshmukh, Ved and Mankikar, Meher and Team, Scale Red and Team, SEAL and Michael, Julian},
  journal={arXiv preprint arXiv:2506.14922},
  year={2025}
}

@article{johnson2011trust,
  title={Trust games: A meta-analysis},
  author={Johnson, Noel D and Mislin, Alexandra A},
  journal={Journal of economic psychology},
  volume={32},
  number={5},
  pages={865--889},
  year={2011},
  publisher={Elsevier}
}

@article{fiske2018stereotype,
  title={Stereotype content: Warmth and competence endure},
  author={Fiske, Susan T},
  journal={Current directions in psychological science},
  volume={27},
  number={2},
  pages={67--73},
  year={2018},
  publisher={Sage Publications Sage CA: Los Angeles, CA}
}

@article{fiske2007universal,
  title={Universal dimensions of social cognition: Warmth and competence},
  author={Fiske, Susan T and Cuddy, Amy JC and Glick, Peter},
  journal={Trends in cognitive sciences},
  volume={11},
  number={2},
  pages={77--83},
  year={2007},
  publisher={Elsevier}
}

@article{jenkins2018predicting,
  title={Predicting human behavior toward members of different social groups},
  author={Jenkins, Adrianna C and Karashchuk, Pierre and Zhu, Lusha and Hsu, Ming},
  journal={Proceedings of the National Academy of Sciences},
  volume={115},
  number={39},
  pages={9696--9701},
  year={2018},
  publisher={National Academy of Sciences}
}

\end{document}